\documentclass[runningheads]{llncs}

\def\cameraready{}

\newif\ifarxivorcameraready
\ifdefined\arxiv
  \arxivorcamerareadytrue
\fi
\ifdefined\cameraready
  \arxivorcamerareadytrue
\fi

\ifdefined\submission
    \usepackage[review,year=2024,ID=1]{eccv}
\fi
\ifarxivorcameraready
    \usepackage[mobile]{eccv}
\fi

\usepackage{eccvabbrv}

\usepackage{graphicx}
\usepackage{booktabs}
\usepackage{todonotes}
\usepackage{subcaption}
\usepackage{adjustbox}

\usepackage[accsupp]{axessibility}  

\ifdefined\submission
    \usepackage[pagebackref,breaklinks,colorlinks,citecolor=eccvblue]{hyperref}
\fi
\ifarxivorcameraready
    \usepackage{hyperref}
\fi

\usepackage{orcidlink}
\usepackage{booktabs}
\usepackage{multirow}
\usepackage{multicol}
\usepackage{wrapfig}
\usepackage{layouts}
\usepackage{float}
\usepackage{array}
\usepackage{tabularx}
\usepackage{xcolor}
\usepackage{tikz}
\usepackage{tcolorbox}
\usepackage{color}
\usepackage{tikz}
\usepackage{soul}

\newcommand{\hlc}[2][yellow]{{%
    \colorlet{foo}{#1}%
    \sethlcolor{foo}\hl{#2}}%
}

\newcommand*\fullcirc[1][1ex]{
  \begin{tikzpicture}
  \fill (0,0) circle (#1);
  \draw[thick] (-#1,0) -- (-4*#1,0);
  \draw[thick] (#1,0) -- (4*#1,0);
  \end{tikzpicture}}

\usepackage{amsmath,amsfonts,bm}
\usepackage{mathtools}

\def\eqref#1{equation~\ref{#1}}

\def\1{\bm{1}}

\DeclareMathAlphabet{\mathsfit}{\encodingdefault}{\sfdefault}{m}{sl}
\SetMathAlphabet{\mathsfit}{bold}{\encodingdefault}{\sfdefault}{bx}{n}

\newcommand*\airemoji{\vcenter{\hbox{\includegraphics[height=\baselineskip]{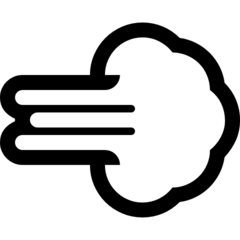}}}}
\newcommand*\mailemoji{\vcenter{\hbox{\includegraphics[height=\baselineskip]{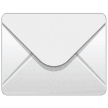}}}}
\newcommand*\oemoji{\vcenter{\hbox{\includegraphics[height=0.8\baselineskip]{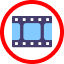}}}}
\newcommand{\acronym}{AirLetters}
\newcommand{\website}{airletters.github.io}

\begin{document}
\crefname{table}{Table}{Tables}
\Crefname{table}{Table}{Tables}
\title{$\mailemoji$~\acronym~$\airemoji$: An Open Vide$\oemoji$ Dataset of Characters Drawn in the Air} 
\titlerunning{\acronym}


\author{Rishit Dagli\inst{\includegraphics[height=0.5\baselineskip]{figures/assets/mailemoji.png}\ \includegraphics[height=0.5\baselineskip]{figures/assets/emoji.png}} \and Guillaume Berger\inst{\includegraphics[height=0.5\baselineskip]{figures/assets/mailemoji.png}} \and Joanna Materzynska\inst{\includegraphics[height=0.5\baselineskip]{figures/assets/oemoji.png}} \and\\ Ingo Bax\inst{\includegraphics[height=0.5\baselineskip]{figures/assets/mailemoji.png}} \and Roland Memisevic\inst{\includegraphics[height=0.5\baselineskip]{figures/assets/mailemoji.png}}}
\authorrunning{R.~Dagli et al.}

\institute{\vspace{-0.7em}$^{\includegraphics[height=0.5\baselineskip]{figures/assets/mailemoji.png}}$\ Qualcomm AI Research\thanks{Qualcomm AI Research is an initiative of Qualcomm Technologies, Inc.}\\$^{\includegraphics[height=0.5\baselineskip]{figures/assets/emoji.png}}$\ University of Toronto\\ $^{\includegraphics[height=0.5\baselineskip]{figures/assets/oemoji.png}}$\ MIT\\
\email{rishit@cs.toronto.edu, guilberg@qti.qualcomm.com, jomat@mit.edu,\\ \{ibax, rmemisev\}@qti.qualcomm.com}}

\maketitle

\definecolor{tgreen}{HTML}{b2df8a}
\definecolor{vblue}{HTML}{1174bc}
\definecolor{tipgreen}{HTML}{33a02c}
\begin{figure}
    \centering
    \includegraphics[]{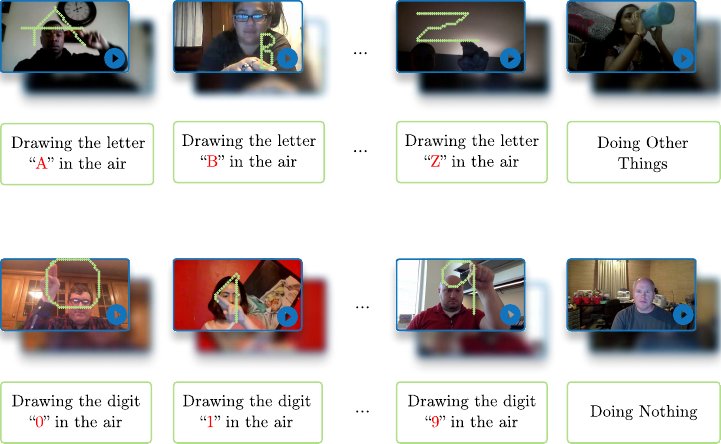}
    \caption{\textbf{Overview.} We present \acronym, a novel dataset comprised of {\color{vblue}video}-{\color{tgreen}label} pairs of human hands denoting \emph{characters} in the \emph{air}. Our dataset contains videos denoting all the Latin \emph{letters} and \emph{digits} as well as two background classes, ``Doing Other Things'' and ``Doing Nothing''. Our dataset contains 161652 videos recorded by 1781 workers. We show the {\color{tipgreen}trajectory} of the fingertips for visualization purposes. 
}
    \label{fig:teaser}
\end{figure}

\vspace{-3em}
\begin{abstract}
We introduce \acronym, a new video dataset consisting of real-world videos of human-generated, articulated motions. Specifically, our dataset requires a vision model to predict letters that humans draw in the air. Unlike existing video datasets, accurate classification predictions for \acronym~rely critically on discerning motion patterns and on integrating long-range information in the video over time. An extensive evaluation of state-of-the-art image and video understanding models on \acronym~shows that these methods perform poorly and fall far behind a human baseline. Our work shows that, despite recent progress in end-to-end video understanding, accurate representations of complex articulated motions -- a task that is trivial for humans -- remains an open problem for end-to-end learning. 
\end{abstract}
\ifdefined\submission
    \newpage
\fi
\section{Introduction}
\label{sec:introduction}

Video understanding is a long-standing research goal in AI. What makes video understanding significantly more challenging than still image understanding is that videos encode information not only spatially but also temporally, in the form of inter-frame correlations, specifically motion. Although the low-level mechanics of extracting spatio-temporal patterns from video are similar to those for extracting spatial patterns from still images (see, for example, \cite{Adelson1985SpatiotemporalEM}), motion understanding relies on visual features that extend across both space and time, and therefore requires operations like 3D convolutions that learn to appropriately aggregate information across these dimensions.

Real-world motion patterns, that
spatio-temporal features learn to represent,
range in complexity from simple transformations, for example, due to camera tilt, to spatio-temporal ``textures'', such as ocean waves and leaves shaking in the wind, to highly complex motion patterns that are generated by articulated (living) bodies. 
Although existing action recognition datasets (such as, 
HMDB-51~\cite{6126543}, UCF-101~\cite{soomro2012ucf101}, ActivityNet-200~\cite{Heilbron_2015_CVPR}, Kinetics~\cite{46330}, Charades~\cite{10.1007/978-3-319-46448-0_31}, TikTokActions~\cite{qian2024advancing},
as well as many others) contain patterns across this spectrum, their labels depend mostly on simple
(across-frame) motion patterns and individual-frame image features. 
For example, to infer a 
label such as ``Baking cookies'' \cite{Heilbron_2015_CVPR} 
with high confidence it suffices to look at a single frame in 
the video. 
As a result, existing video datasets make it hard to learn and evaluate a model's ability to learn complex 
real-world motion patterns.

An exception to this is existing datasets that focus on specific, use-case specific human-generated motion patterns. 
These include, in particular, 
datasets involving hands, which 
can be divided further into video sign language datasets~\cite{forster2012rwth, koller2015continuous, Duarte_CVPR2021, Camgoz_2018_CVPR, Materzynska_2019_ICCV, shi2022opendomain, fang2024signllm} and general hand activity datasets~\cite{Goyal_2017_ICCV}. 
However, since these datasets have been introduced with the task-specific goal of 
understanding sign language, gestures, or hand-object interactions, 
they contain a limited range of  motion patterns, have already saturated performance, 
and in many cases also allow for
inference from individual frames. 



In this work, we introduce \textbf{\acronym}, a novel dataset comprising 161652 labeled videos that capture human hand movements corresponding to digits and letters from the Latin alphabet. 
Our dataset is not only more challenging than existing hand gesture datasets but it also requires models to learn to precisely track hands and analyze long-term dependencies. 
All labels are dynamic and cannot be inferred with one or a few key frames of the video. 
An overview of our dataset is presented in~\Cref{fig:teaser}. 
The dataset also contains two ``contrast classes'' labeled: ``Doing Nothing'' and ``Doing Other Things'', featuring videos of individuals engaged in tasks unrelated to positive labels. 
Due to the in-the-wild nature of the recordings, the videos exhibit considerable variation in lighting conditions, hand positions, backgrounds, drawing motions, and other body movements. These variations render activity recognition within our dataset particularly challenging, necessitating meticulous frame-by-frame analysis. 
Temporal ambiguities (for example, distinguishing between ``O'' and ``Q'' or ``1'' and ``7'') require integration over many frames. Additionally, some common types of data augmentation, such as rotation, are impractical. For example, renderings of the letters ``W'' and ``M'' appear similar under rotation. 
Together these challenges make our dataset a rigorous new testbed for training machine learning models to understand motion in video. 

To showcase the unique challenges and opportunities our dataset presents, we conduct a series of experiments. Through these, we illustrate how our dataset supports the development of models for conventional video understanding and activity recognition. Moreover, the diversity and complexity of the video content in \acronym~makes the dataset useful both as a pre-training dataset and a benchmark in applications in which understanding the motions of human hands is important. 
We also hope that models focusing on video understanding or activity recognition from human hands as well as generative models that focus on generating human hands among others could benefit directly from this dataset\footnote{We plan to make data and code available at  \url{developer.qualcomm.com}.}. 


\section{Related Works}
\label{sec:relatedworks}

Although our data set primarily serves the purpose
of learning and evaluating articulated motion understanding,
it is similar in spirit to
gesture and sign language recognition tasks. 
In this section, we provide a brief overview of existing video sign language benchmarks (\S~\ref{sec:signlanguage}) and video hand gesture benchmarks (\S~\ref{sec:gesture}). We also provide a brief overview of existing general video activity recognition datasets (\S~\ref{sec:actrec}).

\subsection{Sign Language Datasets}
\label{sec:signlanguage}
Historically, the field of video sign language translation has been based on synthetic animation-based methods~\cite{10.1145/638249.638287, KARPOUZIS200754, 10.1145/3490035.3490286, mcdonald2016automated, 10.1145/1531930.1531935}, however, such methods have been replaced by learned approaches~\cite{625742, kadir2004minimal, cooper2007large, koller2015continuous, Cui_2017_CVPR, Camgoz_2018_CVPR, ko2019neural, Camgoz_2020_CVPR, Bohacek_2022_WACV} that require high-quality large-scale data.

A common way to collect sign language data involves crowd-sourcing such the videos and annotations. Many of these datasets contain comprehensive annotations for each gesture in the sign language.  Widely used datasets include CSL-Daily~\cite{Zhou_2021_CVPR} and DEVISIGN~\cite{chai2015devisign} in Chinese Sign Language; KETI~\cite{ko2019neural} in Korean Sign Language; the Public DGS Corpus~\cite{hanke2020extending} in German Sign Language; LSA64~\cite{ronchetti2023lsa64} in Argentinian Sign Language; PSL Kinect 30~\cite{doi:10.5772/60091} and PSL ToF~\cite{doi:10.5772/60091} in Polish Sign Language; GSL~\cite{10.1007/978-3-540-73279-2_73} in Greek Sign Language; and LSE-sign~\cite{gutierrez2016lse} in Spanish Sign Language. These benchmarks feature phrases and dialogues. General word-level American Sign Language datasets include CUNY ASL~\cite{LU2014812}, ASL Lexicon~\cite{4563181}, Purdue RVL-SLLL ASL~\cite{wilbur2006purdue}, and RWTH-BOSTON-50~\cite{zahedi2005combination}, which contain general ASL words but with minimal variance among videos. Other datasets collected this way include How2Sign~\cite{Duarte_CVPR2021}, which features instructional content translated into ASL, and sentence-level datasets like RWTH-BOSTON-104~\cite{zahedi2005combination} and RWTH-BOSTON-400~\cite{zahedi2005combination}. 

Some other large-scale datasets have been taken from television programs with sign language interpreters. These are often limited in variance between videos and usually have some problems with the text alignment. They include datasets like RWTH-PHOENIX-Weather~\cite{8578910} and SWISSTXT~\cite{camgoz2021content4all} which include weather programs in German Sign Language and Swiss German Sign Language, respectively.  
Other datasets derived from television programs include  VRT~\cite{camgoz2021content4all} with news programs in Flemish Sign Language, and BOBSL~\cite{albanie2021bbcoxford} with BBC programs in British Sign Language.

Furthermore, there have also been datasets that are built by scraping videos from the web. Multiple datasets of American Sign Language have been scraped from YouTube like OpenASL~\cite{shi2022opendomain}, and YouTube-ASL~\cite{uthus2023youtubeasl}. There have also been datasets that scrape specialized websites, such as SP-10~\cite{9878501} which includes a multilingual sign language dictionary, and AfriSign~\cite{gueuwou2023afrisign} which translates passages from the Bible, and The Greek Elementary School Dataset~\cite{Voskou_2023_ICCV} with content translated from Greek elementary school content. Lastly, there are many large-scale datasets that scrape videos from the Web, containing:
American Sign Language annotations: MS-ASL~\cite{joze2019msasl}, WLASL~\cite{9093512}, ChicagoFSWild~\cite{8639639}, ChicagoFSWild+~\cite{9010036}, CISLR~\cite{joshi2022cislr}, and Indian-SL~\cite{selvaraj2021openhands}, all of which are word-level datasets; non-ASL annotations: SignsWorld Atlas~\cite{SHOHIEB201568}, LSFB-CONT~\cite{9534336}, LSFB-ISOL~\cite{9534336}, ASL Fingerspelling A~\cite{6130290}, ASL Fingerspelling B~\cite{6130290}, PSL Fingerspelling ToF, Japanese Fingerspelling~\cite{nguen2019deep}, RTWH Fingerspelling~\cite{dreuw2006modeling}, and SIGNUM~\cite{von2007towards}; and multilingual annotations: Prompt2Sign~\cite{fang2024signllm}.

\subsection{Hand Gesture Datasets}
\label{sec:gesture}

The development of datasets in gesture recognition is primarily oriented towards enhancing the precision and versatility of gesture-based interactions in various domains, including human-computer interaction and driving assistance. 
This includes the Cambridge Hand Gesture dataset~\cite{4547427}, which contains 900 RGB sequences across 9 gesture classes, and the Sheffield KInect Gesture (SKIG) dataset~\cite{10.5555/2540128.2540343}, which comprises 1080 RGB-D videos that depict dynamic gestures of 6 participants, categorizing 10 different gestures. 
In parallel, the ChaLearn Gesture Challenge~\cite{Wan_2016_CVPR_Workshops, 10.1145/2522848.2532597} contributed the ChaLearn LAP IsoGD and ConGD datasets~\cite{Wan_2016_CVPR_Workshops}, as well as the Multimodal Gesture Dataset (MMGD)~\cite{10.1145/2522848.2532597}. Some datasets have been captured with sensors, including: MSRGesture3D 2012~\cite{6333871}, ChAirGest 2013~\cite{10.1145/2522848.2532590}, Kinect Numbers and Letters Hand Gestures~\cite{quiroga2013novel}, and LTTM Senz3D~\cite{memo2018head}. 
Some datasets have been captured with imaging equipment. These include Interactive Museum 2014~\cite{Baraldi_2014_CVPR_Workshops}, IPN Hands~\cite{bega2020IPNhand}, LD-ConGR~\cite{ld-congr-cvpr2022}, NUS HandPostures~\cite{10635_56173}, FHANDS~\cite{Garcia-Hernando_2018_CVPR}.
However, most of these datasets are of very small scale or do not have much variance. 
Exceptions are 
Something-Something~\cite{Goyal_2017_ICCV} and Jester~\cite{Materzynska_2019_ICCV}, which are large-scale datasets. 
In contrast to these, our benchmark focuses particularly on specific kinds of gestures that represent Latin characters. Another popular large-scale image dataset is the BIGHands~\cite{Yuan_2017_CVPR} hand pose dataset, which shows significant variance between hand poses but does not represent gestures.

Multiple fine-grained datasets exist for the task of hand gesture recognition. In particular, in the context of automotive applications, datasets such as CVRR-HAND 3D~\cite{6883176} and nvGesture~\cite{Molchanov_2016_CVPR} are specifically designed to understand driver behavior through hand gestures, providing a controlled environment for studying gesture recognition in driving scenarios. Other specialized datasets include  GUN-71~\cite{Rogez_2015_ICCV}, which focuses on fine-grained hand movements for object manipulation, and the NATOPS~\cite{5771448} dataset, which focuses on air signaling gestures for airplanes. 

For first-person perspective applications, datasets such as EgoHands~\cite{Bambach_2015_ICCV}, EgoFinger~\cite{Huang_2016_CVPR_Workshops}, and EgoGesture~\cite{8299578} offer detailed annotations for hand detection and segmentation, capturing data through wearable devices like Google Glass. This perspective is targeted at personal device interactions, and it has been extended to various specialized domains~\cite{karaman2014hierarchical, 7423809}.

\subsection{Activity Recognition and Video Classification Datasets}
\label{sec:actrec}

Video classification and activity recognition involve the categorization of video content into predefined classes. UCF101~\cite{soomro2012ucf101} consists of 13,320 video clips in 101 categories from YouTube, offering diverse and complex activities. HMDB51~\cite{6126543} includes 6,766 video clips across 51 action categories from varied sources like movies and YouTube, presenting challenges such as varying camera angles and lighting. Despite its smaller size, the KTH~\cite{1334462} dataset, with 2,391 video sequences of six actions, laid much of the foundation of early activity recognition research. The Sports-1M~\cite{Karpathy_2014_CVPR} dataset and the Kinetics~\cite{46330} series (Kinetics-400, 600, and 700) are large-scale datasets that have been instrumental in training neural networks for activity recognition tasks. 

Several datasets focus on the fine-grained and contextual understanding of video content. The Charades~\cite{10.1007/978-3-319-46448-0_31} dataset, for instance, focuses on multi-label action recognition through its collection of 9,848 videos depicting everyday indoor activities across 157 action classes. They reflect real-world scenarios where multiple actions coexist. The AVA~\cite{Gu_2018_CVPR} dataset improves fine-grained action recognition by annotating detailed actions within 15-minute movie clips, aiding in spatiotemporal localization.
Hollywood2~\cite{4587756} focuses on actions in realistic settings with videos categorized into 12 human action classes and is used extensively for contextual action recognition. The COIN~\cite{Tang_2019_CVPR} dataset, designed for instructional video analysis, includes 11,827 videos covering 180 tasks in various domains, making it useful for understanding and segmenting instructional content. VideoLT~\cite{zhang2021videolt} tackles the long-tailed distribution problem with its 256,218 untrimmed videos annotated in 1,004 classes, ideal for studying class imbalance. The YouTube-8M~\cite{45619} dataset, comprising 8 million videos annotated with 4,000 visual entities, serves as a large-scale benchmark for video classification models. HVU~\cite{10.1007/978-3-030-58558-7_35} aims to holistically understand videos with 572,000 videos that feature 9 million annotations on 3,142 labels. 
\section{The \acronym\ Dataset}
\label{sec:dataset}

We present the \acronym\ dataset, which is composed of short labeled videos showing people drawing letters in the air with their hands. We next provide details about our video and annotation collection method (\S~\ref{sec:collection}), the content of the dataset (\S~\ref{sec:content}),
and statistics of the dataset (\S~\ref{sec:stats}).

\begin{figure}[tbp]
    \centering
    \begin{tabular}{c}
    Drawing the digit ``2'' in the air \\
    \includegraphics[width=\textwidth]{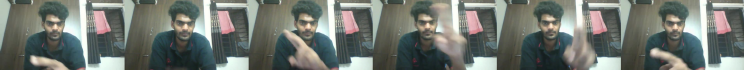} \\
    Drawing the digit ``4'' in the air \\
    \includegraphics[width=\textwidth]{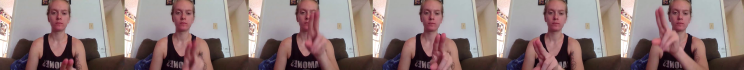} \\
    Drawing the digit ``5'' in the air \\
    \includegraphics[width=\textwidth]{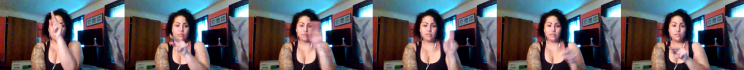} \\
    Drawing the digit ``7'' in the air \\
    \includegraphics[width=\textwidth]{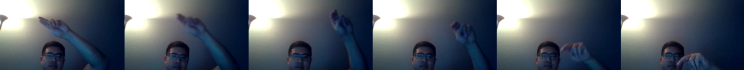} \\
    Drawing the letter ``C'' in the air \\
    \includegraphics[width=\textwidth]{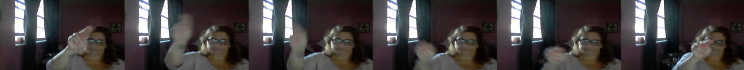} \\
    Drawing the letter ``F'' in the air \\
    \includegraphics[width=\textwidth]{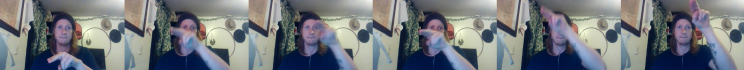} \\
    Drawing the letter ``P'' in the air \\
    \includegraphics[width=\textwidth]{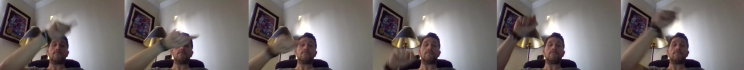} \\
    Drawing the letter ``O'' in the air \\
    \includegraphics[width=\textwidth]{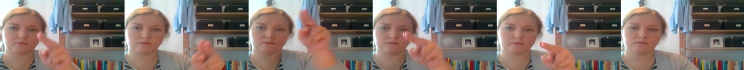} \\
    Doing Other Things \\
    \includegraphics[width=\textwidth]{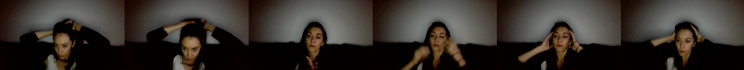} \\
    Doing Nothing \\
    \includegraphics[width=\textwidth]{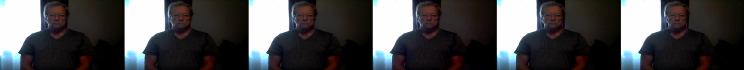}
    \end{tabular}

    \caption{\textbf{Example Videos.} Frames from randomly sampled videos from our dataset showing humans drawing characters as well as contrast classes.}
    \label{fig:examples}
\end{figure}
\begin{figure}[tb]
    \centering
    \begin{tabular}{cc}
        \multirow{1}{*}[3ex]{\rotatebox[origin=c]{90}{B}} & \includegraphics[width=0.97\textwidth]{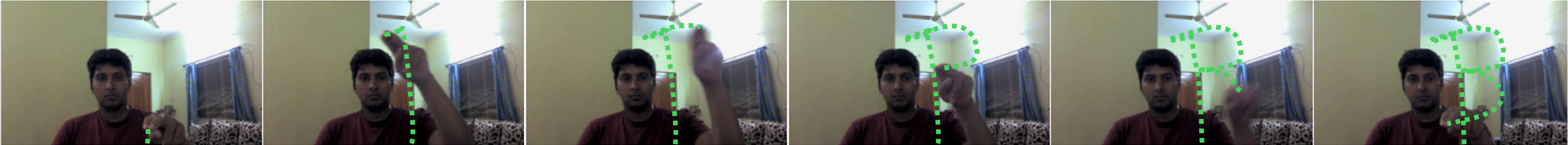}\\
        \multirow{1}{*}[3ex]{\rotatebox[origin=c]{90}{B}} & \includegraphics[width=0.97\textwidth]{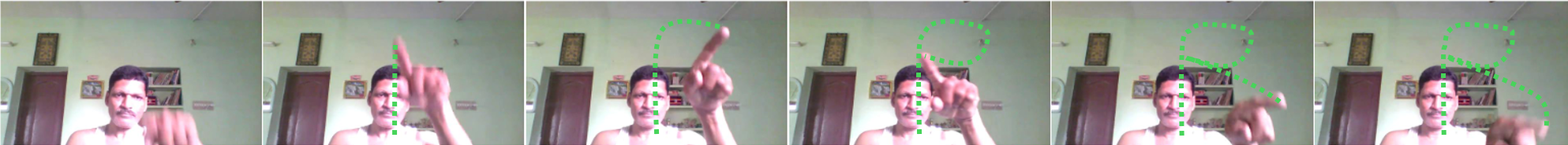}\\
        \multirow{1}{*}[3ex]{\rotatebox[origin=c]{90}{3}} & \includegraphics[width=0.97\textwidth]{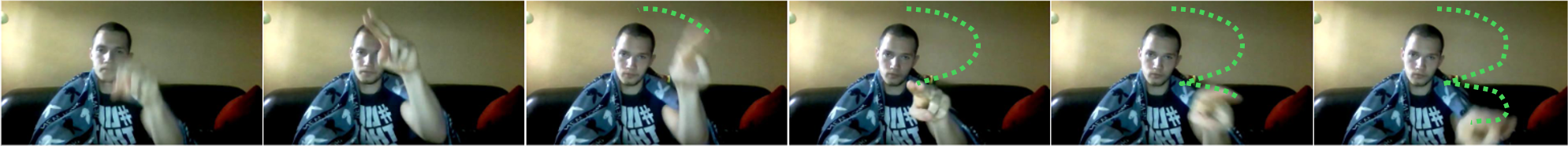}\\
        \multirow{1}{*}[3ex]{\rotatebox[origin=c]{90}{3}} & \includegraphics[width=0.97\textwidth]{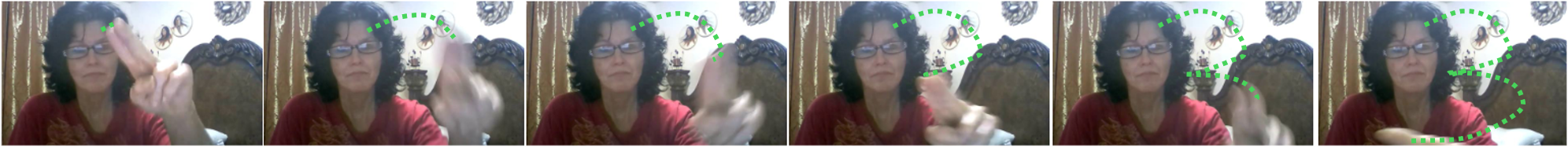}\\
        \midrule
        \multirow{1}{*}[3ex]{\rotatebox[origin=c]{90}{Y}} & \includegraphics[width=0.97\textwidth]{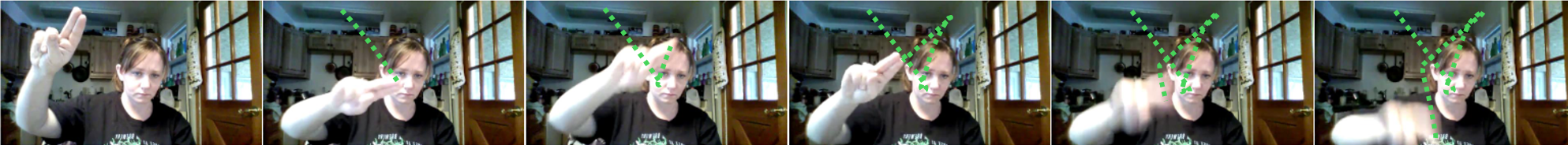}\\
        \multirow{1}{*}[3ex]{\rotatebox[origin=c]{90}{Y}} & \includegraphics[width=0.97\textwidth]{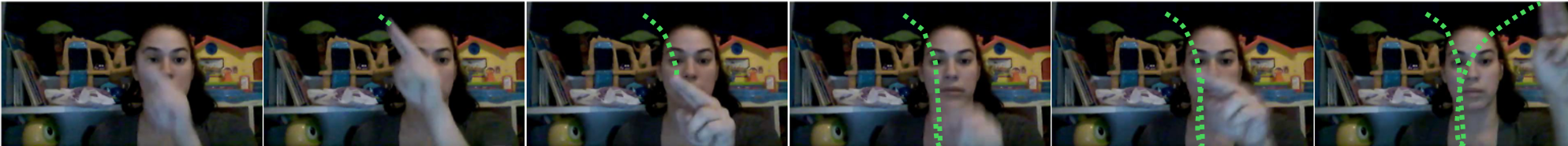}\\
        \multirow{1}{*}[3ex]{\rotatebox[origin=c]{90}{Y}} & \includegraphics[width=0.97\textwidth]{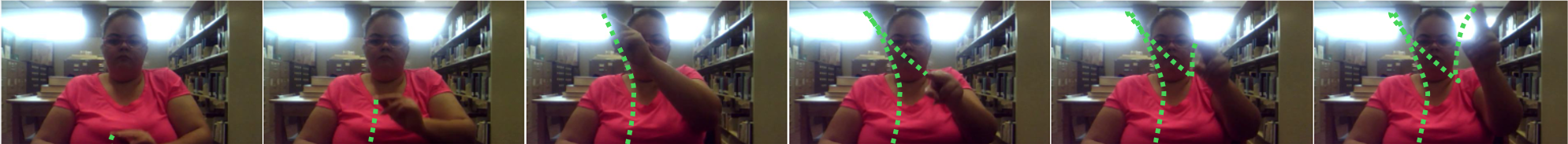}\\
    \end{tabular}
    \caption{\textbf{Challenges due to inter-class similarities and intra-class diversity.} We show some examples of drawing the letter ``B'' and the digit of ``3'', where differentiating both of these classes also requires understanding depth and velocity of relative motion to understand if the individual intended to draw a vertical line (for ``B'') or only meant to place their hands in position (for ``3''). Underneath, we show examples of variability in drawing the letter ``Y''. For example, in one way version of drawing the letter ``Y'', only the last few frames show a stroke that distinguishes it from the letter  ``X''.}
    \label{fig:challenging}
\end{figure}
\begin{figure}[tb]
    \centering
    \includegraphics[width=\textwidth]{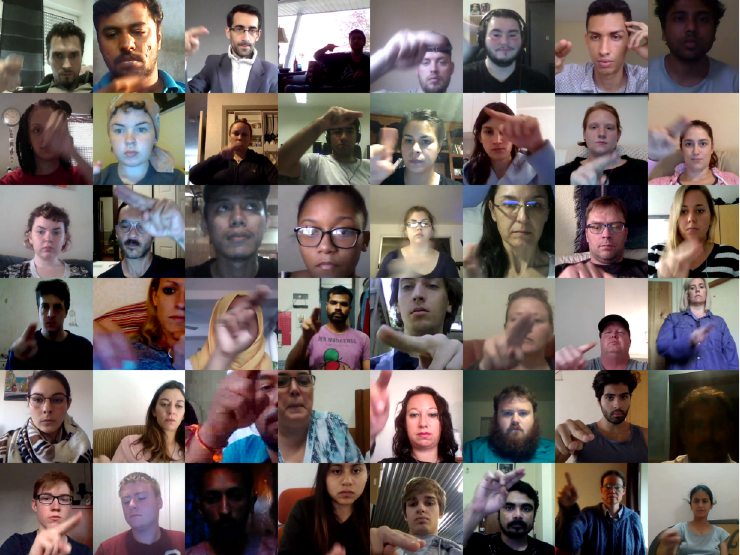}
    \caption{\textbf{Diversity in our Dataset.} Each of the images is taken from a randomly sampled video from our dataset. Our dataset has a large variance in the appearance of subjects, background, occlusion, and lighting conditions in the videos.}
    \label{fig:diversity}
\end{figure}

\subsection{Dataset Content}
\label{sec:content}
The goal of the \acronym\ dataset is to provide a simple,  classification-based evaluation of a model's
ability to correctly understand articulated motions. We focus on manual articulations of each letter of the Latin alphabet as well as numeric digits. This amounts to 36 primary gesture classes, for which recognition requires temporal and spatial analysis of the video. 
The dataset also includes two contrast classes designed to refine the sensitivity and specificity of recognition systems trained on our dataset. The ``Doing Nothing'' class includes videos of individuals in non-active states, such as sitting or standing still, to represent periods of inactivity within human-computer interactions, and the ``Doing Other Things'' class consists of clips capturing miscellaneous, non-communicative movements such as adjusting position or random hand movements. 

We show a few examples from our dataset in~\Cref{fig:examples} using a few
frames per video. 
We also demonstrate the diversity of examples in our dataset in~\Cref{fig:diversity}. Our dataset is curated to reflect real-world complexity, encompassing a range of scenarios where backgrounds are often cluttered and lighting conditions vary from dimly lit to overexposed environments. This heterogeneity 
poses a significant challenge to the robustness of models as they have to deal with a wide spectrum of real-world conditions.

\Cref{fig:challenging} highlights some aspects of our dataset that are challenging for learned models but simple for humans. 
It shows the variability in how participants draw characters, leading to significant variation even within class. 
For example, the letter ``B'' and the digit ``3'' can appear quite distinct depending on the drawing styles of the participants. To accurately differentiate between these two classes, it is essential to analyze the depth and velocity of the relative motion in the videos. This analysis helps determine whether the participant intended to draw a vertical line, indicative of a ``B'', or merely positioned their hands, suggesting a ``3''. Furthermore, we also show the substantial variation in how the letter “Y” is drawn. In some cases, only the final few frames of the drawing process reveal a stroke that is crucial to differentiate ``Y'' from ``X''. 
\setlength{\tabcolsep}{5pt}
\begin{wraptable}{R}{5.5cm}
\vspace{-0.45cm}
\caption{\textbf{Dataset Splits.} The number of crowd workers and videos in each split of our dataset.}
\label{tab:splits}
\centering
\begin{tabular}{crr}\\\toprule  
Split & Videos & Workers \\\midrule
Train & 128745 & 958\\
Validation & 16480 & 412\\
Test & 16427 & 411\\ \bottomrule
\end{tabular}
\vspace{-1.2cm}
\end{wraptable}

We roughly split the dataset using an 8:1:1 ratio for training, validation, and testing,  respectively. To do so, we assign each one of the 1781 crowd workers to either the training, validation, or test split. We show the number of videos in each split in~\Cref{tab:splits}. 

\subsection{Dataset Statistics}
\label{sec:stats}

Our dataset is designed to mirror real-world conditions and showcases a diverse range of backgrounds and variations, with 1781 crowd workers contributing. It consists of 161652 videos, with each contributor recording an average of 90.76 videos at an average frame rate of 30 frames per second (fps) to accommodate different recording devices. The average duration of each video is approximately 2.92 seconds, allowing for the completion of the required gesture without prolonging the recording unnecessarily. 
The spatial resolution of the videos has an 
area averaging 0.25 megapixels at varying aspect ratios. 
The total number of frames per video varies depending on the frame rate, but on average, each video contains $\approx$252 frames. We show summaries of these statistics in~\Cref{tab:stats}.

\subsection{Collection Methodology}
\label{sec:collection}

To collect our dataset, we used a custom platform integrated with crowd-sourcing providers. 
This allowed us to recruit participants from diverse gender, geographical, and ethnic backgrounds and to provide the required instruction and recording functionality.
Participants redirected to our platform were asked to record themselves performing all 36 gestures in front of their camera. We provided detailed visual and textual instructions to ensure clear hand visibility, high video quality, and precise gesture execution. Supplementary example videos were provided to demonstrate correct gestures and to address the limitations of text instructions. After reviewing the guidelines, participants prepared for recording with the help of a countdown timer. Recordings averaged $\approx$3 seconds, after which participants could review and re-record if necessary. 
For added variability, the “Doing Other Things” category required four distinct activities, while the “Doing Nothing” category required no specific activities. Each participant could make up to three submissions. 
To encourage scene variability, participants could interrupt recording and resume at a later time.

All submissions were reviewed by human operators to verify accuracy. Participants with mostly correct submissions but minor errors were allowed to make corrections and resubmit. This approach ensured the high quality and consistency of the dataset. Finally, all videos were resized to a width of 640 pixels, maintaining the aspect ratio.


\begin{table}[tb]
    \caption{\textbf{Dataset Statistics,} showing the number of classes, number of actors and median values for duration, frames per second (FPS), videos per class, and videos per actor.}
    \label{tab:stats}
    \begin{subtable}{.4\linewidth}
      \centering
        \begin{tabular}{lr}
          \toprule
          \textbf{Statistic} & \textbf{Value (Total)}\\
          \midrule
          Videos & 161652\\
          Classes & 38\\
          Actors & 1781\\
          Frames & 40142100\\
          \bottomrule
        \end{tabular}
    \end{subtable}%
    \begin{subtable}{.6\linewidth}
      \centering
        \begin{tabular}{lr}
          \toprule
          \textbf{Statistic} & \textbf{Value (Median, $\sigma$)}\\
          \midrule
          Duration & 2.93 {\textbf{\scriptsize\textcolor{gray}{($\pm$0.13)}}}\\
          FPS & 30.0 {\textbf{\scriptsize\textcolor{gray}{($\pm$0.0)}}}\\
          Videos per Class ($\times10^3$) & 4.04 {\textbf{\scriptsize\textcolor{gray}{($\pm$1.31)}}}\\
          Videos per Actor & 40.0 {\textbf{\scriptsize\textcolor{gray}{($\pm$99.29)}}}\\
          \bottomrule
        \end{tabular}
    \end{subtable} 
\end{table}
\section{Experiments Validating \acronym}
\label{sec:experiments}

We conduct various experiments to assess the difficulty of this task. Below, we highlight the baseline architectures we used (\S~\ref{sec:models}), our preprocessing workflow (\S~\ref{sec:preprocess}), and present our results (\S~\ref{sec:results}).

\subsection{Baseline Architectures}
\label{sec:models}

\definecolor{tabfirst}{rgb}{1, 0.7, 0.7} 
\definecolor{tabsecond}{rgb}{1, 0.85, 0.7} 
\definecolor{tabthird}{rgb}{1, 1, 0.7} 
\begin{table}[tb]
    \caption{\textbf{Classification accuracy} of multiple image models, video models, and (large) vision language models on the  \acronym~dataset. Note that the task is straightforward for humans but challenging for existing models.}
    \label{tab:results}
    \centering
    \begin{tabular}{lr}
         \toprule
         \textbf{Method} & \textbf{Top-1 Accuracy ($\uparrow$)} \\
         \midrule
         \emph{Image Models} & \\
         \midrule
         ViT-B/16~\cite{dosovitskiy2021an} & 7.49 \\
         MaxViT-T~\cite{10.1007/978-3-031-20053-3_27} & 7.56 \\
         ResNet-200~\cite{he2015deepresiduallearningimage} & 11.44 \\
         ResNeXt-101~\cite{Xie_2017_CVPR} & 13.09 \\
         SE-ResNeXt-26~\cite{rw2019timm} & 13.29 \\
         ResNet-50~\cite{he2015deepresiduallearningimage} & 13.87 \\
         \midrule
         \emph{Video Models} & \\
         \midrule
         VideoMAE (16)~\cite{NEURIPS2022_416f9cb3} & 57.96\\
         ResNet-101 + LSTM & 58.45\\
         ResNet-50 + LSTM & 63.24\\
         ResNext-152 3D & 65.77\\
         Strided Inflated EfficientNet 3D~\cite{mercier2023end} & 65.97\\
         ResNext-50 3D & \cellcolor{tabthird}66.54\\
         ResNext-101 3D & \cellcolor{tabsecond}69.74\\
         ResNext-200 3D & \cellcolor{tabfirst}71.20\\
         \midrule
         \emph{Vision Language Models} & \\
         \midrule
         Video-LLaVA (w/o contrast class)~\cite{lin2023videollava} & 2.53\\
         VideoLLaMA2 (w/o contrast class)~\cite{cheng2024videollama2advancingspatialtemporal} & 2.47\\
         Video-LLaVA~\cite{lin2023videollava} & 7.29\\
         VideoLLaMA2~\cite{cheng2024videollama2advancingspatialtemporal} & 7.58\\
         \midrule
         Human Performance (10 videos/class) & 96.67\\
         \bottomrule
    \end{tabular}
\end{table}
\begin{figure}[tb]
    \centering
    \includegraphics[width=\textwidth]{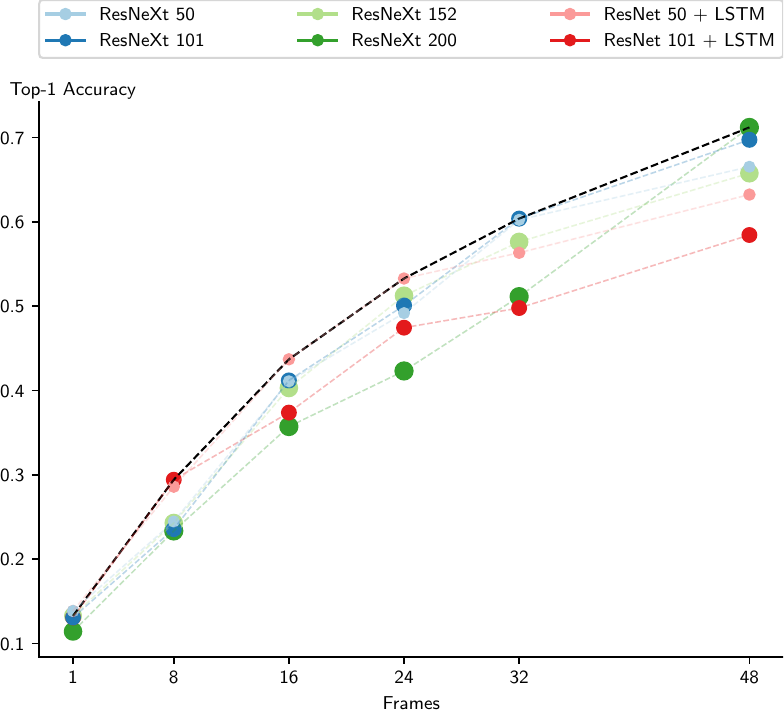}
    \caption{\textbf{Scaling Training Frames.} Performance of models across different numbers of training frames. The Pareto Frontier is represented by a black curve (\fullcirc[0.75ex]). Note  that this dataset requires models to attend through the entire video to perform well, and increasing the number of frames that models attend to significantly increases their performance.}
    \label{fig:scaling}
\end{figure}

\paragraph{Image Models.} We train baseline image classification models, including ResNet~\cite{he2015deepresiduallearningimage}, ResNeXt~\cite{Xie_2017_CVPR}, SE ResNeXt~\cite{rw2019timm}, MaxViT~\cite{10.1007/978-3-031-20053-3_27}, and ViT~\cite{dosovitskiy2021an} to predict the activity label given a single video frame. During testing, we average model outputs for each frame of the test videos to produce a final prediction.

\paragraph{Video Models.} We also train baseline video models, including ResNet 3D, ResNeXt 3D, Strided Inflated EfficientNet 3D~\cite{mercier2023end}, and VideoMAE~\cite{NEURIPS2022_416f9cb3}. 
Since the data are inherently temporal, we
also train a ResNet~\cite{he2015deepresiduallearningimage} baseline paired with an LSTM~\cite{6795963}, where the ResNet backbone extracts 2D features from individual frames and the features are passed to an LSTM layer. We use the last hidden state as the encoding for the videos. 
We compare training from scratch, fine-tuning from models pre-trained on either Kinetics~\cite{46330} or ImageNet~\cite{5206848}, as well as finetuning from  Imagenet pre-trained classifiers whose parameters are inflated to 3D~\cite{8099985}. 

\paragraph{Vision Language Models.} We also experiment with identifying actions from videos in a zero-shot manner using a large vision language model, specifically Video LLaVa~\cite{lin2023videollava} and Video Llama2~\cite{cheng2024videollama2advancingspatialtemporal}. 
This includes experiments, where we remove the two contrast classes we have while evaluating these models to demonstrate the difficulty they have in estimating the non-contrast classes.

We train the baseline models using either Adam~\cite{kingma2017adammethodstochasticoptimization} or AdamW~\cite{loshchilov2018decoupled} and adopt the standard cross-entropy loss with label smoothing~\cite{Szegedy_2016_CVPR}.
Depending on the model, we experiment with various learning rates and schedules, including constant learning rates, cosine decay, and exponential decay. We present experimental details in~\Cref{sec:experimentaldetails}.

\subsection{Preprocessing Workflow}
\label{sec:preprocess}

Before training, we resample all of our videos to 30 FPS and resize the videos in an aspect-ratio preserving manner to 300 pixels for the shortest edge.
We use a standard video processing workflow during training and testing. We sample the videos with FPS $\in\left[8, 24\right]$. In the case that sampling at 8 FPS does not leave us with at least the number of frames required for the model, we shift the lower bound to an FPS that can give us at least the number of frames needed. We then perform a spatio-temporal crop on the videos.
During evaluation and testing, we perform a center crop, followed by sampling the required number of frames by performing a temporal center crop.

\begin{figure}[tb!]
    \centering
    \includegraphics[width=\textwidth]{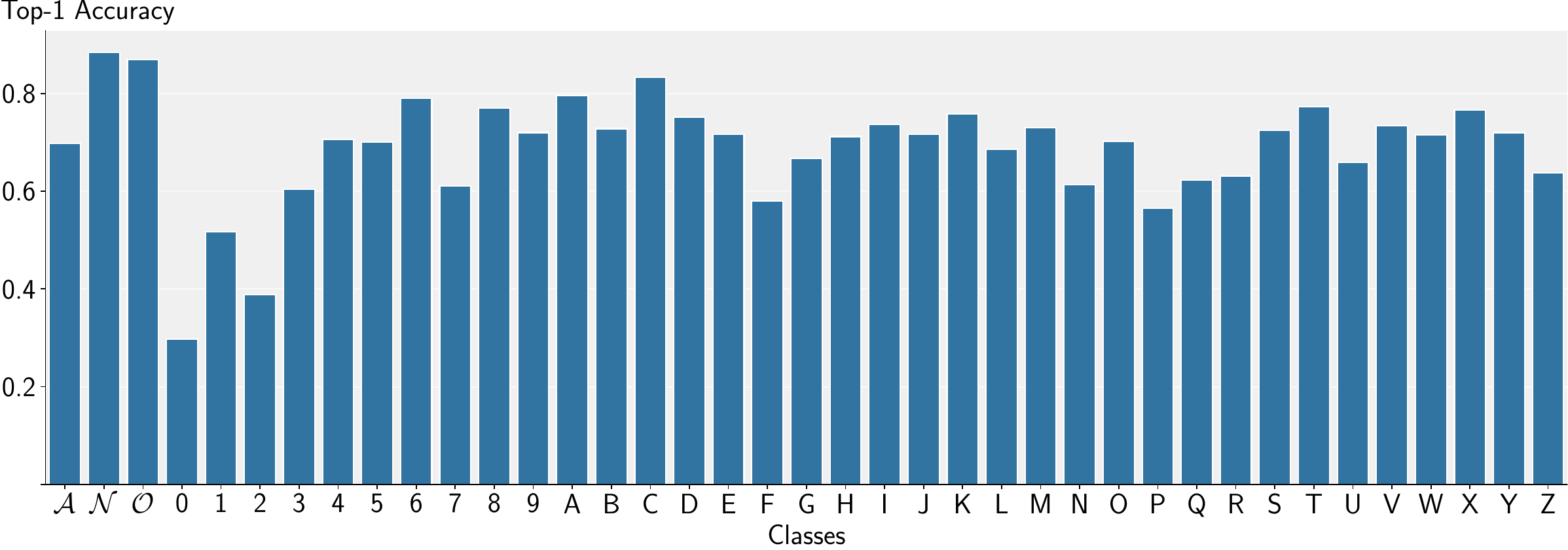}
    \caption{\textbf{Top-1 accuracy for each class} for the best-performing model from~\Cref{tab:results}, where $\mathcal{A}$ represents the average top-1 accuracy, $\mathcal{N}$ the class ``Doing Nothing'' and $\mathcal{O}$ the class ``Doing Other Things''.}
    \label{fig:classacc}
\end{figure}
\begin{figure}[h!]
    \centering
    \includegraphics[width=\textwidth]{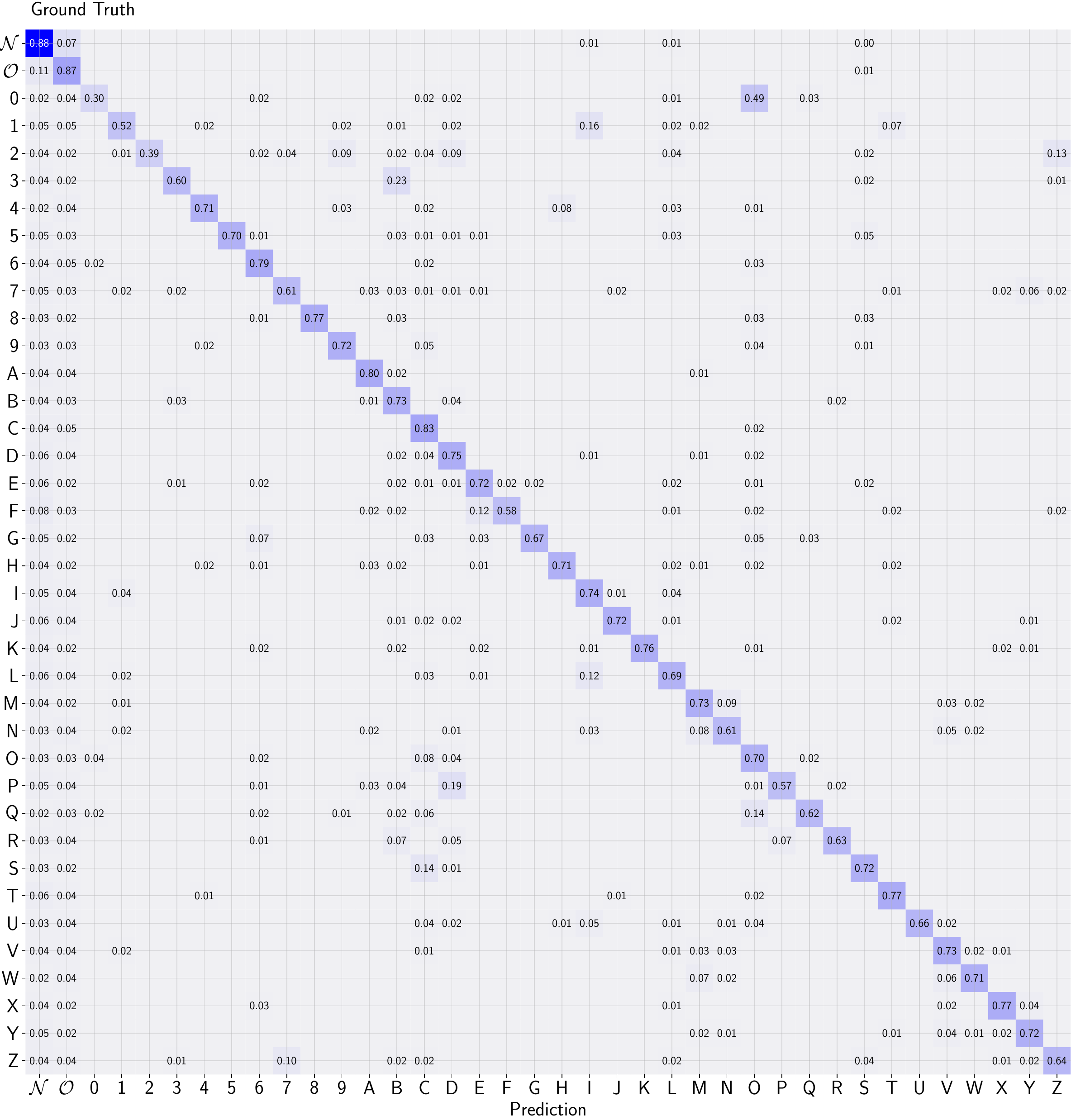}
    \caption{\textbf{Confusion Matrix} for the best-performing model from~\Cref{tab:results}, where $\mathcal{N}$ represents the class ``Doing Nothing'' and $\mathcal{O}$ represents the class ``Doing Other Things''. 
    }
    \label{fig:confusion}
\end{figure}

\subsection{Results}
\label{sec:results}

We evaluated the baseline architectures described in Section \S~\ref{sec:models} for the task of end-to-end video activity recognition on our data set, and report the top-1 accuracy in~\Cref{tab:results}.
Our results highlight a significant gap in current end-to-end video understanding and activity recognition methods: all models, especially large vision language models, perform well below human evaluation results. Human evaluation achieves near-perfect accuracy, while the task is challenging for all tested models. 

We examine the performance of models trained on different numbers of frames: 1 (image models), 8, 16, 24, 32, and 48 frames per video in~\Cref{fig:scaling}. We observe a significant increase in the performance of models when they are trained on more frames.  
On average, the videos are sampled at 16 FPS (due to our pre-processing described in~\Cref{sec:preprocess}) and have a duration of approximately seconds (\Cref{sec:dataset}). 
We notice a significant increase in performance when increasing the number of frames from 32 to 48, demonstrating that our dataset requires models to attend to most frames of the video to perform well on this benchmark. Furthermore, our experiments also validate that our dataset requires models to learn long-range temporal dependencies and to have the ability to aggregate information temporally.

We also show the top-1 accuracy for each of the classes of a ResNeXt-3D model in~\Cref{fig:classacc} and the corresponding confusion matrix in~\Cref{fig:confusion}. We observe that classes such as the digits ``0'', ``1'', and ``2'' are particularly challenging, as they are easily confused with each other. In contrast, the contrast classes ``Doing Nothing'' and ``Doing Other Things'', are more easily recognized. 
We also notice some expected misclassification patterns in~\Cref{fig:confusion}, such as ``0'' and ``O'', ``3'' and ``B'', or ``P'' and ``D'' being misclassified for one another due to the visual similarity of these characters.

\section{Conclusion}
\label{sec:conclusion}
We introduced a new real-world dataset, 
utilizing human generated articulated motions. 
Unlike existing video datasets, accurate predictions for our dataset require detailed
understanding of motion and the integration of
long-range information across the video. 
We show that existing image and video understanding models perform poorly and fall
far behind the human baseline. 
In this way, our work exposes a significant gap in the current video understanding capabilities. 
Closing this gap, arguably, will be a 
necessary step to build AI models that can perceive the world more like humans. 

\newpage
\bibliographystyle{splncs04}
\bibliography{main}

\newpage
\appendix

\definecolor{darkblue}{RGB}{86, 65, 150}
\newcommand{\dsquestion}[1]{%
    {\noindent \textcolor{darkblue}{\textbf{#1}}}
}
\newcommand{\dsquestionex}[2]{%
    {\noindent \textcolor{darkblue}{\textbf{#1} #2}}
}
\newcommand{\dsanswer}[1]{%
   {\noindent #1 \medskip}
}

\section{Experimental Details}
\label{sec:experimentaldetails}

We show the prompts we used to evaluate the vision language models in~\Cref{fig:model_input}.

All of our experiments use PyTorch 1.13~\cite{NEURIPS2019_bdbca288} and Accelerate~\cite{accelerate} to train our models in a distributed fashion. Our codebase is based on \texttt{timm}~\cite{rw2019timm} and HuggingFace Transformers~\cite{Wolf_Transformers_State-of-the-Art_Natural_2020}. We present all the experimental details for training the models from~\Cref{tab:results} in~\Cref{tab:videodetails1,tab:videodetails2,tab:imagedetails}, for all other hyper-parameters we use PyTorch defaults.

 \begin{table}[!b]
    \centering
    \caption{\textbf{Experimental Details for Video Models I}, whose performance is shown in \Cref{tab:results}.}
    \label{tab:videodetails1}
    \begin{adjustbox}{max width=\textwidth}
    \begin{tabular}{l|r|r}
    \toprule
    \multicolumn{1}{l}{Model} & \multicolumn{1}{r}{ResNeXt-200 3D} & \multicolumn{1}{r}{Strided Inflated}\\
    \multicolumn{1}{r}{} & \multicolumn{1}{r}{} & \multicolumn{1}{r}{EfficientNet 3D~\cite{mercier2023end}}\\
    \midrule
    Training Precision & FP-32 & FP-32\\
    \# of frames & 48 & 48\\
    Frame Size & $(224, 224)$ & $(224, 224)$\\
    Initialization & ImageNet-1k & Properiatery Dataset\\
                   & w/ RA1 recipe~\cite{wightman2021resnet} & w/ recipe~\cite{mercier2023end}\\
    Training Preprocessing & Random Resized & Random Resized\\
                           & Crop, $(0.7,1.0)$ & Crop, $(0.7,1.0)$\\
    Eval Preprocessing & Center Crop & Center Crop\\
    Label Smoothing & $10^{-1}$ & $10^{-1}$\\
    Batch Size & 32 & 32\\
    Optimizer & AdamW~\cite{loshchilov2018decoupled} & AdamW~\cite{loshchilov2018decoupled}\\
    Optimizer Parameters & $\lambda = 10^{-2}$ & $\lambda = 10^{-2}$\\
                         & $\beta_1 = 0.9$ & $\beta_1 = 0.9$\\
                         & $\beta_2 = 0.999$ & $\beta_2 = 0.999$\\
    Initial learning rate & $10^{-4}$ & $10^{-3}$\\
    LR Schedule & Static & Static\\
    Scheduler Parameters & &\\
    Gradient clipping & None & None\\
    Training Iterations & 385k & 385k\\
    Params (M) & 67.77 & 14.46\\
    \bottomrule
    \end{tabular}
    \end{adjustbox}
\end{table}
    
\begin{table}[tb]
    \centering
    \caption{\textbf{Experimental Details for Video Models II,} whose performance is shown in \Cref{tab:results}.}
    \label{tab:videodetails2}
    \begin{adjustbox}{max width=\textwidth}
    \begin{tabular}{l|r|r|r}
    \toprule
    \multicolumn{1}{l}{Model} & \multicolumn{1}{r}{VideoMAE~\cite{NEURIPS2022_416f9cb3}} & \multicolumn{1}{r}{ResNet-101 + LSTM} & \multicolumn{1}{r}{ResNet-50 + LSTM}\\
    \midrule
    Training Precision & FP-32 & FP-32 & FP-32\\
    \# of frames & 16 & 48 & 48\\
    Frame Size & $(224, 224)$ & $(224, 224)$ & $(224, 224)$\\
    Initialization & ImageNet-1k & ImageNet-1k & ImageNet-1k\\
                   & w/ MAE~\cite{MaskedAutoencoders2021} & w/ RA1 recipe~\cite{wightman2021resnet} & w/ RA1 recipe~\cite{wightman2021resnet}\\
    Training Preprocessing & Random Resized & Random Resized & Random Resized\\
                           & Crop, $(0.7,1.0)$ & Crop, $(0.7,1.0)$ & Crop, $(0.7,1.0)$\\
    Eval Preprocessing & Center Crop & Center Crop & Center Crop\\
    Label Smoothing & $10^{-1}$ & $10^{-1}$ & $10^{-1}$\\
    Batch Size & 8 & 8 & 32\\
    Optimizer & AdamW~\cite{loshchilov2018decoupled} & Adam~\cite{kingma2017adammethodstochasticoptimization} & Adam~\cite{kingma2017adammethodstochasticoptimization}\\
    Optimizer Parameters & $\lambda = 10^{-2}$ & $\lambda = 0$ & $\lambda = 0$\\
                         & $\beta_1 = 0.9$ & $\beta_1 = 0.9$ & $\beta_1 = 0.9$\\
                         & $\beta_2 = 0.999$ & $\beta_2 = 0.999$ & $\beta_2 = 0.999$\\
    Initial learning rate & $10^{-5}$ & $10^{-4}$ & $10^{-3}$\\
    LR Schedule & Cosine Annealing & Static & Static\\
                & w/ Warm Restart & & \\
    Scheduler Parameters & $T_0 = 2$ & &\\
                         & $\eta_{\text{min}} = 10^{-2}$ & &\\
                         & $\alpha_{max} = 10^{-2}$ & &\\
    Gradient clipping & None & None & None\\
    Training Iterations & 600k & 385k & 385k\\
    Params (M) & 86.26 & 43.72 & 24.73\\
    \bottomrule
    \end{tabular}
    \end{adjustbox}
    \begin{adjustbox}{max width=\textwidth}
    \begin{tabular}{l|r|r|r}
    \toprule
    \multicolumn{1}{l}{Model} & \multicolumn{1}{r}{Resnext-50 3D} & \multicolumn{1}{r}{Resnext-101 3D} & \multicolumn{1}{r}{Resnext-152 3D}\\
    \midrule
    Training Precision & FP-32 & FP-32 & FP-32\\
    \# of frames & 48 & 48 & 48\\
    Frame Size & $(224, 224)$ & $(224, 224)$ & $(224, 224)$\\
    Initialization & ImageNet-1k & ImageNet-1k & ImageNet-1k\\
                   & w/ RA1 recipe~\cite{wightman2021resnet} & w/ RA1 recipe~\cite{wightman2021resnet} & w/ RA1 recipe~\cite{wightman2021resnet}\\
    Training Preprocessing & Random Resized & Random Resized & Random Resized\\
                           & Crop, $(0.7,1.0)$ & Crop, $(0.7,1.0)$ & Crop, $(0.7,1.0)$\\
    Eval Preprocessing & Center Crop & Center Crop & Center Crop\\
    Label Smoothing & $10^{-1}$ & $10^{-1}$ & $10^{-1}$\\
    Batch Size & 32 & 32 & 64\\
    Optimizer & Adam~\cite{kingma2017adammethodstochasticoptimization} & Adam~\cite{kingma2017adammethodstochasticoptimization} & AdamW~\cite{loshchilov2018decoupled}\\
    Optimizer Parameters & $\lambda = 0$ & $\lambda = 0$ & $\lambda = 10^{-2}$\\
                         & $\beta_1 = 0.9$ & $\beta_1 = 0.9$ & $\beta_1 = 0.9$\\
                         & $\beta_2 = 0.999$ & $\beta_2 = 0.999$ & $\beta_2 = 0.999$\\
    Initial learning rate & $10^{-4}$ & $10^{-4}$ & $10^{-4}$\\
    LR Schedule & Static & Static & Static\\
    Scheduler Parameters &&&\\
    Gradient clipping & None & None & None\\
    Training Iterations & 385k & 385k & 385k\\
    Params (M) & 23.17 & 44.82 & 62.66\\
    \bottomrule
    \end{tabular}
    \end{adjustbox}
\end{table}

\begin{table}[tb]
    \centering
    \caption{\textbf{Experimental Details for Image Models,} whose performance is shown in \Cref{tab:results}.}
    \label{tab:imagedetails}
    \begin{adjustbox}{max width=\textwidth}
    \begin{tabular}{l|r|r|r}
    \toprule
    \multicolumn{1}{l}{Model} & \multicolumn{1}{r}{ViT-B/16~\cite{dosovitskiy2021an}} & \multicolumn{1}{r}{MaxViT-T~\cite{10.1007/978-3-031-20053-3_27}} & \multicolumn{1}{r}{ResNet-200~\cite{he2015deepresiduallearningimage}}\\
    \midrule
    Training Precision & FP-32 & FP-32 & FP-32\\
    Frame Size & $(224, 224)$ & $(224, 224)$ & $(224, 224)$\\
    Initialization & ImageNet-1k & ImageNet-1k & ImageNet-1k\\
                   & w/ MAE~\cite{MaskedAutoencoders2021} & (TF Weights) & w/ RA2 recipe~\cite{wightman2021resnet}\\
    Training Preprocessing & Center Crop & Center Crop & Center Crop\\
    Eval Preprocessing & Center Crop & Center Crop & Center Crop\\
    Label Smoothing & $10^{-1}$ & $10^{-1}$ & $10^{-1}$\\
    Batch Size & 512 & 64 & 512\\
    Optimizer & AdamW~\cite{loshchilov2018decoupled} & AdamW~\cite{loshchilov2018decoupled} & AdamW~\cite{loshchilov2018decoupled}\\
    Optimizer Parameters & $\lambda = 5\times10^{-2}$ & $\lambda = 5\times10^{-2}$ & $\lambda = 5\times10^{-2}$\\
                         & $\beta_1 = 0.9$ & $\beta_1 = 0.9$ & $\beta_1 = 0.9$\\
                         & $\beta_2 = 0.999$ & $\beta_2 = 0.999$ & $\beta_2 = 0.999$\\
    Initial learning rate & $10^{-5}$ & $10^{-3}$ & $10^{-3}$\\
    LR Schedule & Cosine Annealing & Cosine Annealing & Cosine Annealing\\
                & w/ Warm Restart & w/ Warm Restart & w/ Warm Restart\\
    Scheduler Parameters & $T_0 = 2$ & $T_0 = 2$ & $T_0 = 2$\\
                         & $\eta_{\text{min}} = 10^{-2}$ & $\eta_{\text{min}} = 10^{-2}$ & $\eta_{\text{min}} = 10^{-2}$\\
                         & $\alpha_{max} = 10^{-2}$ & $\alpha_{max} = 10^{-2}$ & $\alpha_{max} = 10^{-2}$\\
    Gradient clipping & None & None & None\\
    Training Iterations & 230k & 230k & 180k\\
    Params (M) & 85.83 & 28.56 & 62.72\\
    \bottomrule
    \end{tabular}
    \end{adjustbox}
    \begin{adjustbox}{max width=\textwidth}
    \begin{tabular}{l|r|r|r}
    \toprule
    \multicolumn{1}{l}{Model} & \multicolumn{1}{r}{ResNeXt-101~\cite{Xie_2017_CVPR}} & \multicolumn{1}{r}{SE ResNeXt~\cite{rw2019timm}} & \multicolumn{1}{r}{ResNet-50~\cite{he2015deepresiduallearningimage}}\\
    \midrule
    Training Precision & FP-32 & FP-32 & FP-32\\
    Frame Size & $(224, 224)$ & $(224, 224)$ & $(224, 224)$\\
    Initialization & ImageNet-1k & YFCC100M & ImageNet-1k\\
                   & w/ MAE~\cite{MaskedAutoencoders2021} & FT ImageNet-1k & w/ RA1 recipe~\cite{wightman2021resnet}\\
    Training Preprocessing & Center Crop & Center Crop & Center Crop\\
    Eval Preprocessing & Center Crop & Center Crop & Center Crop\\
    Label Smoothing & $10^{-1}$ & $10^{-1}$ & $10^{-1}$\\
    Batch Size & 1024 & 1024 & 1024\\
    Optimizer & AdamW~\cite{loshchilov2018decoupled} & AdamW~\cite{loshchilov2018decoupled} & AdamW~\cite{loshchilov2018decoupled}\\
    Optimizer Parameters & $\lambda = 5\times10^{-2}$ & $\lambda = 5\times10^{-2}$ & $\lambda = 5\times10^{-2}$\\
                         & $\beta_1 = 0.9$ & $\beta_1 = 0.9$ & $\beta_1 = 0.9$\\
                         & $\beta_2 = 0.999$ & $\beta_2 = 0.999$ & $\beta_2 = 0.999$\\
    Initial learning rate & $10^{-4}$ & $10^{-4}$ & $10^{-3}$\\
    LR Schedule & Cosine Annealing & Cosine Annealing & Cosine Annealing\\
                & w/ Warm Restart & w/ Warm Restart & w/ Warm Restart\\
    Scheduler Parameters & $T_0 = 2$ & $T_0 = 2$ & $T_0 = 2$\\
                         & $\eta_{\text{min}} = 10^{-2}$ & $\eta_{\text{min}} = 10^{-2}$ & $\eta_{\text{min}} = 10^{-2}$\\
                         & $\alpha_{max} = 10^{-2}$ & $\alpha_{max} = 10^{-2}$ & $\alpha_{max} = 10^{-2}$\\
    Gradient clipping & None & None & None\\
    Training Iterations & 180k & 180k & 180k\\
    Params (M) & 42.58 & 14.84 & 23.59\\
    \bottomrule
    \end{tabular}
    \end{adjustbox}
\end{table}



\section{Additional Examples}

\begin{figure}[tb]
    \centering
    \begin{tcolorbox}[colframe=blue, colback=white, title=Video-LLaVA, fonttitle=\bfseries, width=\textwidth]
        \textbf{Q:} \texttt{USER}: \texttt{<video>}You are a video classifier trained to detect letters and digits drawn by humans in the air with their fingers. You will be provided with a video of a person drawing in the air, carefully analyze the video to determine what letter or digit the person draws. Only respond with the character detected, no explanation. The only valid responses are "doing other things" or "doing nothing" or a letter or a digit. \texttt{ASSISTANT}:\\\\
        \textbf{A:} \hlc[cyan!50]{doing other things}
    \end{tcolorbox}
    \begin{tcolorbox}[colframe=blue, colback=white, title=Video-LLaVA (w/o  contrast classes), fonttitle=\bfseries, width=\textwidth]
        \textbf{Q:} \texttt{USER}: \texttt{<video>}You are a video classifier trained to detect letters and digits drawn by humans in the air with their fingers. You will be provided with a video of a person drawing in the air, carefully analyze the video to determine what letter or digit the person draws. Only respond with the character detected, no explanation. The only valid responses are a letter or a digit. \texttt{ASSISTANT}:\\\\
        \textbf{A:} \hlc[cyan!50]{A}
    \end{tcolorbox}
    \begin{tcolorbox}[colframe=blue, colback=white, title=VideoLLaMA2, fonttitle=\bfseries, width=\textwidth]
        \textbf{Q:} You are a video classifier trained to detect letters and digits drawn by humans in the air with their fingers. You will be provided with a video of a person drawing in the air, carefully analyze the video to determine what letter or digit the person draws. Only respond with the character detected, no explanation. The only valid responses are "doing other things" or "doing nothing" or a letter or a digit.\\\\
        \textbf{A:} \hlc[cyan!50]{doing nothing}
    \end{tcolorbox}
    \begin{tcolorbox}[colframe=blue, colback=white, title=VideoLLaMA2 (w/o  contrast classes), fonttitle=\bfseries, width=\textwidth]
        \textbf{Q:} You are a video classifier trained to detect letters and digits drawn by humans in the air with their fingers. You will be provided with a video of a person drawing in the air, carefully analyze the video to determine what letter or digit the person draws. Only respond with the character detected, no explanation. The only valid responses are "doing other things" or "doing nothing" or a letter or a digit.\\\\
        \textbf{A:} \hlc[cyan!50]{2}
    \end{tcolorbox}
    \caption{\textbf{Vision Language Model Evaluation.} We show the prompts we use to evaluate vision-language models.}
    \label{fig:model_input}
\end{figure}

\begin{figure}
    \centering
    \includegraphics[width=\textwidth]{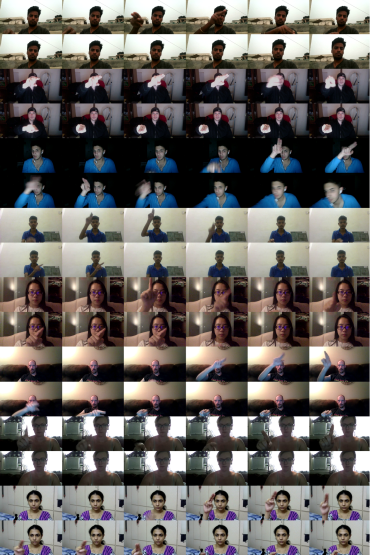}
    \caption{\textbf{Additional Examples.} We show a few more video examples from our dataset by sampling 12 frames uniformly from randomly selected videos.}
    \label{fig:additionalexamples}
\end{figure}

We demonstrate additional video examples from our dataset in~\Cref{fig:additionalexamples}. We also show a histogram showing the distribution of videos among classes. All classes have an equal number of videos except ``doing other things'' which has 3$\times$ the number of videos. 

\begin{figure}[tb]
    \centering
    \includegraphics[width=\textwidth]{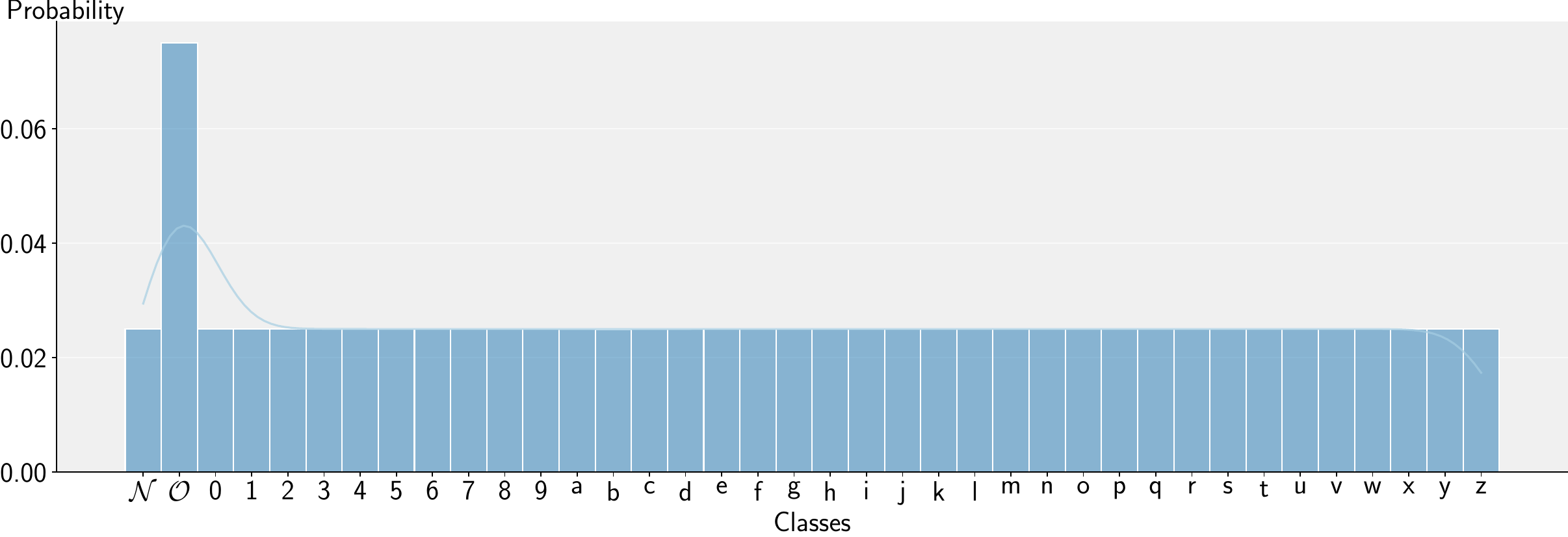}
    \caption{\textbf{Distribution over classes in the \acronym~dataset}.}
    \label{fig:classes-hist}
\end{figure}

\section{Datasheet}

We present a datasheet for our dataset inspired by the template in~\cite{gebru2021datasheets}.

\subsection{Motivation}

\dsquestion{For what purpose was the dataset created?}

\dsanswer{The purpose of the \acronym\ dataset is to support training and evaluation of the ability of models to recognize motion patterns and to perform temporal aggregation of information across a video.
}

\dsquestion{Who created this dataset (e.g., which team, research group) and on behalf of which entity (e.g., company, institution, organization)?}

\ifarxivorcameraready
    \dsanswer{The dataset was created by the authors of the paper on behalf of Qualcomm Technologies Inc. and TwentyBN GmbH.
    }
\fi
\ifdefined\submission
    \dsanswer{We skip this question to comply with the double-blind review.
    }
\fi

\dsquestion{Who funded the creation of the dataset?}

\ifarxivorcameraready
    \dsanswer{The creation of this dataset was funded by Qualcomm Technologies Inc. and TwentyBN GmbH.
    }
\fi
\ifdefined\submission
    \dsanswer{We skip this question to comply with the double-blind review.
    }
\fi

\dsquestion{Any other comments?}

\dsanswer{No.
}

\subsection{Composition}

\dsquestionex{What do the instances that comprise the dataset represent (e.g., documents, photos, people, countries)?}{ Are there multiple types of instances (e.g., movies, users, and ratings; people and interactions between them; nodes and edges)? Please provide a description.}

\dsanswer{An instance of the dataset consists of a video of a person drawing a character in the air with their hands, as well as a corresponding label.
}

\dsquestion{How many instances are there in total (of each type, if appropriate)?}

\dsanswer{There are 161652 videos in total. Each of the classes has an equal number of videos except for the class ``doing other things'' which has 3$\times$ the number of videos.
}

\dsquestionex{Does the dataset contain all possible instances or is it a sample (not necessarily random) of instances from a larger set?}{ If the dataset is a sample, then what is the larger set? Is the sample representative of the larger set (e.g., geographic coverage)? If so, please describe how this representativeness was validated/verified. If it is not representative of the larger set, please describe why not (e.g., to cover a more diverse range of instances, because instances were withheld or unavailable).}

\dsanswer{This dataset contains all possible instances and is not a sample from a larger set.}

\dsquestionex{What data does each instance consist of? “Raw” data (e.g., unprocessed text or images) or features?}{In either case, please provide a description.}

\dsanswer{An instance of the dataset consists of a video as well as the following information:
\begin{description}
    \item[Worker ID.] A unique integer worker ID to differentiate between individuals who recorded the videos.
    \item[Duration.] Duration of the video.
    \item[Label.] The letter or digit drawn by the worker using their hand(s) or one of the contrast classes: doing nothing or doing other things. 
\end{description}
}

\dsquestionex{Is there a label or target associated with each instance?}{If so, please provide a description.}

\dsanswer{A label can be a letter, a digit, or one of the contrast classes: ``doing nothing'' or ``doing other things''.
}

\dsquestionex{Is any information missing from individual instances?}{If so, please provide a description, explaining why this information is missing (e.g., because it was unavailable). This does not include intentionally removed information, but might include, e.g., redacted text.}

\dsanswer{No.}

\dsquestionex{Are relationships between individual instances made explicit (e.g., users’ movie ratings, social network links)?}{If so, please describe how these relationships are made explicit.}

\dsanswer{N/A.
}

\dsquestionex{Are there recommended data splits (e.g., training, development/validation, testing)?}{If so, please provide a description of these splits, explaining the rationale behind them.}

\dsanswer{The dataset has an 8:1:1 split into training set, validation set and test set.
}

\dsquestionex{Are there any errors, sources of noise, or redundancies in the dataset?}{If so, please provide a description.}

\dsanswer{Videos were reviewed to detect potential errors, but it is not guaranteed from being free of any errors, noise or redundancies.}

\dsquestionex{Is the dataset self-contained, or does it link to or otherwise rely on external resources (e.g., websites, tweets, other datasets)?}{If it links to or relies on external resources, a) are there guarantees that they will exist, and remain constant, over time; b) are there official archival versions of the complete dataset (i.e., including the external resources as they existed at the time the dataset was created); c) are there any restrictions (e.g., licenses, fees) associated with any of the external resources that might apply to a future user? Please provide descriptions of all external resources and any restrictions associated with them, as well as links or other access points, as appropriate.}

\dsanswer{The dataset is self-contained.
}

\dsquestionex{Does the dataset contain data that might be considered confidential (e.g., data that is protected by legal privilege or by doctor-patient confidentiality, data that includes the content of individuals non-public communications)?}{If so, please provide a description.}

\dsanswer{No.}

\dsquestionex{Does the dataset contain data that, if viewed directly, might be offensive, insulting, threatening, or might otherwise cause anxiety?}{If so, please describe why.}

\dsanswer{No.
}

\dsquestionex{Does the dataset relate to people?}{If not, you may skip the remaining questions in this section.}

\dsanswer{Yes, the dataset contains videos of humans drawing characters in the air with their hands.
}

\dsquestionex{Does the dataset identify any subpopulations (e.g., by age, gender)?}{If so, please describe how these subpopulations are identified and provide a description of their respective distributions within the dataset.}

\dsanswer{No.}

\dsquestionex{Is it possible to identify individuals (i.e., one or more natural persons), either directly or indirectly (i.e., in combination with other data) from the dataset?}{If so, please describe how.}

\dsanswer{
While the faces of the individuals in the video are visible, the videos were collected under a direct agreement with the crowd workers, permitting research and commercial use. The audio and meta-data information from the videos was removed.
}

\dsquestionex{Does the dataset contain data that might be considered sensitive in any way (e.g., data that reveals racial or ethnic origins, sexual orientations, religious beliefs, political opinions or union memberships, or locations; financial or health data; biometric or genetic data; forms of government identification, such as social security numbers; criminal history)?}{If so, please provide a description.}

\dsanswer{No, this dataset does not contain sensitive data.
}

\dsquestion{Any other comments?}

\dsanswer{No.
}

\subsection{Collection Process}

\dsquestionex{How was the data associated with each instance acquired?}{Was the data directly observable (e.g., raw text, movie ratings), reported by subjects (e.g., survey responses), or indirectly inferred/derived from other data (e.g., part-of-speech tags, model-based guesses for age or language)? If data was reported by subjects or indirectly inferred/derived from other data, was the data validated/verified? If so, please describe how.}

\dsanswer{The dataset was collected with the help of crowdworkers and contractors.}


\dsanswer{A simple web interface was used for recording videos and creating annotations. The resulting data was manually inspected to ensure data integrity.
}

\dsquestion{If the dataset is a sample from a larger set, what was the sampling strategy (e.g., deterministic, probabilistic with specific sampling probabilities)?}

\dsanswer{N/A}



\dsquestionex{Over what timeframe was the data collected? Does this timeframe match the creation timeframe of the data associated with the instances (e.g., recent crawl of old news articles)?}{If not, please describe the timeframe in which the data associated with the instances was created.}

\dsanswer{N/A.}



\dsquestionex{Does the dataset relate to people?}{If not, you may skip the remaining questions in this section.}

\dsanswer{Yes, the dataset contains videos of humans drawing characters in the air with their hands.
}

\dsquestion{Did you collect the data from the individuals in question directly, or obtain it via third parties or other sources (e.g., websites)?}

\dsanswer{The data was collected directly. 
}

\dsquestionex{Were the individuals in question notified about the data collection?}{If so, please describe (or show with screenshots or other information) how notice was provided, and provide a link or other access point to, or otherwise reproduce, the exact language of the notification itself.}

\dsanswer{Yes.}

\dsquestionex{Did the individuals in question consent to the collection and use of their data?}{If so, please describe (or show with screenshots or other information) how consent was requested and provided, and provide a link or other access point to, or otherwise reproduce, the exact language to which the individuals consented.}

\dsanswer{Yes, crowdworkers signed a consent form.
}

\dsquestionex{If consent was obtained, were the consenting individuals provided with a mechanism to revoke their consent in the future or for certain uses?}{If so, please provide a description, as well as a link or other access point to the mechanism (if appropriate).}

\ifarxivorcameraready
    \dsanswer{Yes, the participants may reach out to us via email:
    
    \href{mailto:research.datasets@qti.qualcomm.com}{\texttt{research.datasets@qti.qualcomm.com}}.
    }
\fi
\ifdefined\submission
    \dsanswer{Yes, the participants may reach out to us via email.
    }
\fi





\subsection{Preprocessing/cleaning/labeling}

\dsquestionex{Was any preprocessing/cleaning/labeling of the data done (e.g., discretization or bucketing, tokenization, part-of-speech tagging, SIFT feature extraction, removal of instances, processing of missing values)?}{If so, please provide a description. If not, you may skip the remainder of the questions in this section.}

\dsanswer{Based on the aspect ratio of the originally recorded video, we perform an aspect-ratio preserving resizing to the width off 640 pixels. As a result, all videos are either (360, 640) or (480, 640). All of videos are preprocessed to have a framerate of 30 FPS.
}

\dsquestionex{Was the “raw” data saved in addition to the preprocessed/cleaned /labeled data (e.g., to support unanticipated future uses)?}{If so, please provide a link or other access point to the “raw” data.}

\dsanswer{No.
}

\dsquestionex{Is the software used to preprocess/clean/label the instances available?}{If so, please provide a link or other access point.}

\dsanswer{No.
}



\subsection{Uses}

\dsquestionex{Has the dataset been used for any tasks already?}{If so, please provide a description.}

\dsanswer{Yes. In this work, baseline models are evaluated on the dataset.
}

\dsquestionex{Is there a repository that links to any or all papers or systems that use the dataset?}{If so, please provide a link or other access point.}

\dsanswer{Currently, such a repository does not exist. 
}

\dsquestion{What (other) tasks could the dataset be used for?}

\dsanswer{The dataset can be also be used for pre-training video understanding or video generation models. 
}







\subsection{Distribution}

\dsquestionex{Will the dataset be distributed to third parties outside of the entity (e.g., company, institution, organization) on behalf of which the dataset was created?}{If so, please provide a description.}

\ifarxivorcameraready
    \dsanswer{Yes, we plan to make the dataset publicly available.
    }
\fi
\ifdefined\submission
    \dsanswer{Yes, we plan to make the dataset publicly available.
    }
\fi

\dsquestionex{How will the dataset be distributed (e.g., tarball on website, API, GitHub)}{Does the dataset have a digital object identifier (DOI)?}

\dsanswer{The dataset will be publicly downloadable through a website.
}

\dsquestion{When will the dataset be distributed?}

\ifarxivorcameraready
    \dsanswer{The dataset should be available publicly on the dataset website.
    }
\fi
\ifdefined\submission
    \dsanswer{The dataset should be publicly available prior to ECCV 2024.
}
\fi

\dsquestionex{Will the dataset be distributed under a copyright or other intellectual property (IP) license, and/or under applicable terms of use (ToU)?}{If so, please describe this license and/or ToU, and provide a link or other access point to, or otherwise reproduce, any relevant licensing terms or ToU, as well as any fees associated with these restrictions.}

\dsanswer{Yes, we plan to release the dataset under a proprietary research license.}

\dsquestionex{Have any third parties imposed IP-based or other restrictions on the data associated with the instances?}{If so, please describe these restrictions, and provide a link or other access point to, or otherwise reproduce, any relevant licensing terms, as well as any fees associated with these restrictions.}

\dsanswer{No.
}

\dsquestionex{Do any export controls or other regulatory restrictions apply to the dataset or to individual instances?}{If so, please describe these restrictions, and provide a link or other access point to, or otherwise reproduce, any supporting documentation.}

\dsanswer{No.
}



\subsection{Maintenance}

\dsquestion{Who will be supporting/hosting/maintaining the dataset?}

\ifarxivorcameraready
    \dsanswer{The dataset is hosted and maintained by Qualcomm Technologies Inc.
    }
\fi
\ifdefined\submission
    \dsanswer{We skip this question to comply with the double-blind review.
    }
\fi

\dsquestion{How can the owner/curator/manager of the dataset be contacted (e.g., email address)?}

\ifarxivorcameraready
    \dsanswer{The owners of the dataset can be contacted through:
    
    \href{mailto:research.datasets@qti.qualcomm.com}{\texttt{research.datasets@qti.qualcomm.com}}.
    }
\fi
\ifdefined\submission
    \dsanswer{We skip this question to comply with the double-blind review.
    }
\fi

\dsquestionex{Is there an erratum?}{If so, please provide a link or other access point.}

\dsanswer{N/A
}

\dsquestionex{Will the dataset be updated (e.g., to correct labeling errors, add new instances, delete instances)?}{If so, please describe how often, by whom, and how updates will be communicated to users (e.g., mailing list, GitHub)?}

\dsanswer{If the dataset is updated, changes should be communicated through the dataset web page.
}

\dsquestionex{If the dataset relates to people, are there applicable limits on the retention of the data associated with the instances (e.g., were individuals in question told that their data would be retained for a fixed period of time and then deleted)?}{If so, please describe these limits and explain how they will be enforced.}

\dsanswer{No.
}

\dsquestionex{Will older versions of the dataset continue to be supported/hosted/ maintained?}{If so, please describe how. If not, please describe how its obsolescence will be communicated to users.}

\dsanswer{N/A.}

\dsquestionex{If others want to extend/augment/build on/contribute to the dataset, is there a mechanism for them to do so?}{If so, please provide a description. Will these contributions be validated/verified? If so, please describe how. If not, why not? Is there a process for communicating/distributing these contributions to other users? If so, please provide a description.}

\dsanswer{Certain mechanisms exists for research use cases. Further information is detailed in the proprietary research license.
}



\end{document}